\documentclass[twoside,twocolumn]{article}

\usepackage[T1]{fontenc}
\usepackage[utf8]{inputenc}

\usepackage{blindtext} 

\usepackage[sc]{mathpazo} 
\usepackage[T1]{fontenc} 
\usepackage{microtype} 

\usepackage[english]{babel} 
\usepackage{hyperref}
\usepackage{graphicx}
\usepackage{float}
\usepackage{listings}
\usepackage{color,soul}
\usepackage{comment}
\usepackage{relsize}
\definecolor{mygreen}{rgb}{0,0.6,0}
\definecolor{mygray}{rgb}{0.5,0.5,0.5}
\definecolor{mymauve}{rgb}{0.58,0,0.82}
\usepackage[hmarginratio=1:1,top=32mm,columnsep=20pt,margin=0.8in]{geometry} 
\usepackage[hang, small,labelfont=bf,up,textfont=it,up]{caption}
\usepackage[hang, small,labelfont=bf,up,textfont=it,up]{subcaption} 
\usepackage{booktabs} 
 
\usepackage{lettrine} 

\usepackage{enumitem} 
\setlist[itemize]{noitemsep} 

\usepackage{abstract} 

\usepackage{titlesec} 
\renewcommand\thesection{\Roman{section}} 
\renewcommand\thesubsection{\roman{subsection}} 
\titleformat{\section}[block]{\large\scshape\centering}{\thesection.}{1em}{} 
\titleformat{\subsection}[block]{\large}{\thesubsection.}{1em}{} 

\usepackage{fancyhdr} 
\usepackage{titling} 

\usepackage{hyperref} 

\pretitle{\begin{center}\Huge\bfseries} 
\posttitle{\end{center}} 
\title{Closing the Gap in Swarm Robotics Simulations: An Extended Ardupilot/Gazebo plugin} 
\author{%
\textsc{Hugo R. M. Sardinha}  \\[1ex]
\normalsize Edinburgh Centre for Robotics\\ 
\normalsize \href{mailto:hs20@hw.ac.uk}{hs20@hw.ac.uk} 
\and 
\textsc{Mauro Dragone} \\[1ex] 
\normalsize Edinburgh Centre for Robotics\\
\normalsize \href{m.dragone@hw.ac.uk}{m.dragone@hw.ac.uk}
\and
\textsc{Patricia A. Vargas} \\[1ex] 
\normalsize Edinburgh Centre for Robotics \\
\normalsize \href{mailto:
p.a.vargas@hw.ac.uk}{
p.a.vargas@hw.ac.uk}
}
\date{\today} 
\renewcommand{\maketitlehookd}{%
\begin{abstract}
\noindent 
This work provides an extension of Ardupilot's capabilities to allow researchers to develop swarm robotics applications in Robot Operating System(ROS)/Gazebo simulations. Ardupilot is a tool used for autopilot functions in autonomous air, land and underwater robots. However, the original Ardupilot was not designed for applications in the field of swarm robotics. Hence, the main contribution of this paper is to endow Ardupilot with the capability to handle swarm robotics simulations, including the refactoring of the manufacturer's description files. This work offers a comprehensive and realistic simulation test-bed for swarm robotics applications under the ROS/Ardupilot umbrella using an existing off-the-shelf Erlerobotics' erlecopter UAV.

\end{abstract}
}

\begin{document}

\maketitle


\section{Introduction}

\begin{figure}[H]
\centering
\includegraphics[width=0.9\linewidth]{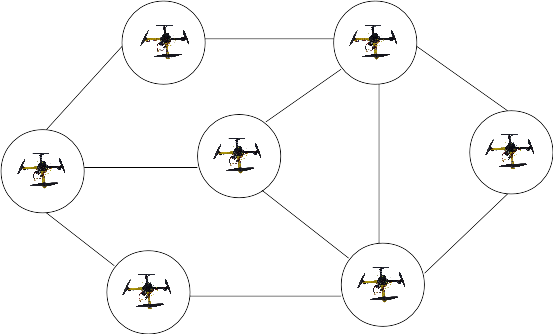}
\caption{A concept of the connectivity in a UAV swarm}
\label{fig1}
\end{figure}

For the past decades, autonomous vehicles have been the subject of intensive research both in academia and industry \cite{yan2017building} \cite{couceiro2014benchmark} \cite{faigl2015benchmarking}. Amongst those, the Unmanned Aerial Vehicle (UAV) has become a major platform to study swarm robotics applications due to its maneuverability and lower costs. However, despite the substantial increase in research in this area, a platform that enables the realization of swarm simulations with UAVs in a widespread manner is still not available. 
\\
Across the literature, the major issues with swarm simulations is that the platforms are either very simple, or sophisticated but proprietary. In the first case, the UAV possess almost no sensing beyond a gyroscope, and most of the remaining sensing is done from external sources, such as tracking systems. On the other hand, proprietary platforms make it very hard for researchers to do any benchmarking study. Replicating results is often impossible as teams do not have access to the same hardware and/or software. \noindent Hence, a freely accessible platform that facilitates the benchmarking of results and approaches, leveraging local connectivity, would be greatly beneficial. 
\\
The need for a comprehensive framework in swarm simulation is clearly demonstrated by the increasing interest, not only in swarm applications, but also in their comparative performances, as pointed out in some benchmark studies \cite{couceiro2014benchmark} \cite{faigl2015benchmarking}. Figure \ref{fig1} shows a concept of connectivity in a UAV swarm.

This work's main contribution is to enable the development of UAV swarm applications in a very realistic simulation environment \cite{koenig2004use} using Ardupilot. Ardupilot is a tool used for autopilot functions in autonomous air, land and underwater robots. Ardupilot is already used in single-UAV research \cite{andrade2017autonomous} but not ready for swarm applications. Hence, the outcomes of this paper are of great relevance to the future of the field as it addresses a clear gap in the swarm robotics research.

\noindent 



\section{Related Work}

Recent efforts have been made in designing a suitable framework for multi-robot systems simulations \cite{yan2017building}. The authors of this work focus on building a ROS-compatible framework and use MORSE (\textbf{M}odular \textbf{O}pen\textbf{R}obots \textbf{S}imulation \textbf{E}ngine) simulator to achieve a realistic simulation, arguing that their framework would be simulator independent, such as Stage \cite{gerkey2003player} or Gazebo \cite{koenig2004use}.
%
%
The work presented here pushes the realism of the simulation yet another step further by incorporating Ardupilot 
into a ROS/Gazebo environment. The ardupilot software is an autopilot software 
which provides a wide range of functionalities such as sensor checks, takeoff and land commands, as well as way-point navigation. By providing these tools, ardupilot is also, on a simulation level, meant to narrow the reality gap. 
Consequently, Ardupilot is already used in single-UAV research \cite{andrade2017autonomous}, a fact which indicates that Ardupilot-powered vehicles may become a tendency in the future, hence justifying its improvement to swarm applications.\\

\vspace{-10pt}
\section{Proposed Software Architecture}
The proposed plugin and its related files can be found on the following the links\footnotemark \footnotetext{\href{https://github.com/Hurisa/ardupilot\_sitl\_gazebo\_plugin}{https://github.com/Hurisa/ardupilot\_sitl\_gazebo\_plugin}} \footnotemark \footnotetext {\href{https://github.com/Hurisa/multi\_uav/blob/master/worlds/empty.world}{https://github.com/Hurisa\\/multi\_uav/blob/master/worlds/empty.world}}\\

\begin{figure}[!htb]
\centering
\includegraphics[width=0.45\textwidth]{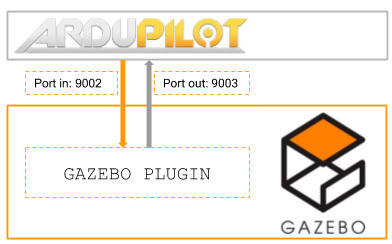}
\caption{Generic Communication architecture between ardupilot and gazebo}
\label{fig:arch1}
\end{figure}

This section shows the constraints of Ardupilot related to swarm systems and how Ardupilot and Gazebo work together. 
The foremost aspect one must understand is that this communication architecture between ardupilot and gazebo is made via a Gazebo plugin.
When Ardupilot is launched, two ports are automatically open for input and output with the external software. These ports are numbered \textbf{9002} for input, and \textbf{9003} for output. As a result, when the plugin is launched it expects these two ports will be open and establishes a connection. Figure \ref{fig:arch1} illustrates this architecture.



\begin{figure}[htp]
\centering
\includegraphics[width=0.45\textwidth]{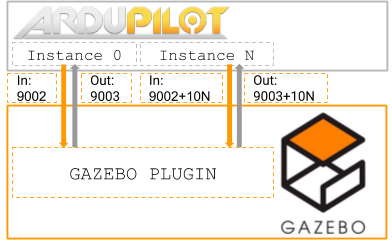}
\caption{Proposed Communication architecture between ardupilot and gazebo}
\label{fig:arch2}
\end{figure}

In the original version of the plugin, no port number is specified when the plugin is called and the connection between Ardupilot and Gazebo is limited to two ports. Nevertheless, as new instances of Ardupilot are initiated it creates port numbers of the type \textbf{9002+10*n} and \textbf{9003+10*n}, for input and output respectively, where \textbf{n} is the instance number, starting at zero. 
The bottleneck that prevents the simulation of several robots resides on the plugin itself. Since the plugin always expects to connect to the default ports, a connection will never be established to new instances of Ardupilot, resulting in conflicting vehicles attempting to connect to the same ports. 
\noindent The key concept of this work is to fork the original plugin and change it so as to pass the port numbers as arguments. Since the plugin is launched in Gazebo it receives no information of what port numbers were created by the Ardupilot instantiations, thus when the new instance of the plugin is called, the corresponding port numbers must be indicated. Figure \ref{fig:arch2} shows the proposed architecture. 
%

By changing the plugin's architecture to receive user-specified port numbers, it enabled the connection of ardupilot to several simulated robots.

\section{Case Study}

This section will show the results of an implementation of a flocking behaviour using the framework that was developed. It will serve as a proof of concept to demonstrate how the changes made to the Ardupilot/Gazebo plugin enabled the development of swarm techniques in a scalable manner. 

Flocking is one of the most studied behaviours in swarms and has been implemented in both simulated \cite{moeslinger2010emergent},and real robots \cite{turgut2008self}. Therefore our contribution will leverage these works by facilitating the development of realistic simulations that could be easily disseminated and benchmarked. Reynolds in \cite{reynolds1987flocks} established the three main rules for simulated flocks, which can be summarized as follows:

\begin{itemize}
\item \textbf{Collision Avoidance} - A minimum distance is maintained from the individuals in the neighbourhood
\item \textbf{Velocity Matching} - velocities are aligned with those of the neighbourhood
\item \textbf{Flock Centering} - velocities are also weighted by center of neighbourhood and change towards its center.
\end{itemize}

This work assess the flocking behaviour by employing the metrics suggested by Turgut and collaborators in\cite{turgut2008self}, namely:

\begin{itemize}
\item \textit{Order} ($\psi$) - This parameter measures the angular variance of robots using the expression 
\begin{equation}
\psi(t)=\frac{1}{N} \left| \sum_{k=1}^{N} e^{i\theta_k}\right|
\label{eq:oreder}
\end{equation}
Where $N$ is the number of robots in the swarm and $\theta_k$ is the orientation of robot $k$ at time $t$. This parameter can be viewed online by recording the attitude of each robot at each control step. If the swarm is aligned the value $\psi$ is closer to 1, while on the other hand if the swarm is not aligned, $\psi$
is closer to 0.\\

\item \textit{Swarm Velocity} ($V_s$): This corresponds to the average velocity value of the swarm during the course of the experiment. A high swarm velocity indicates an efficient and smooth motion whereas a low swarm velocity indicates inefficiency. This metric is expressed as follows:
\begin{equation}
V_s=\frac{ \mathlarger{\sum_{i=1}^{N}}\left|v_i\right|}{N}
\label{eq:swarmvel}
\end{equation}
\end{itemize}
\section{Platform}
\begin{figure}[!htp]
\centering
\includegraphics[width=0.3\textwidth]{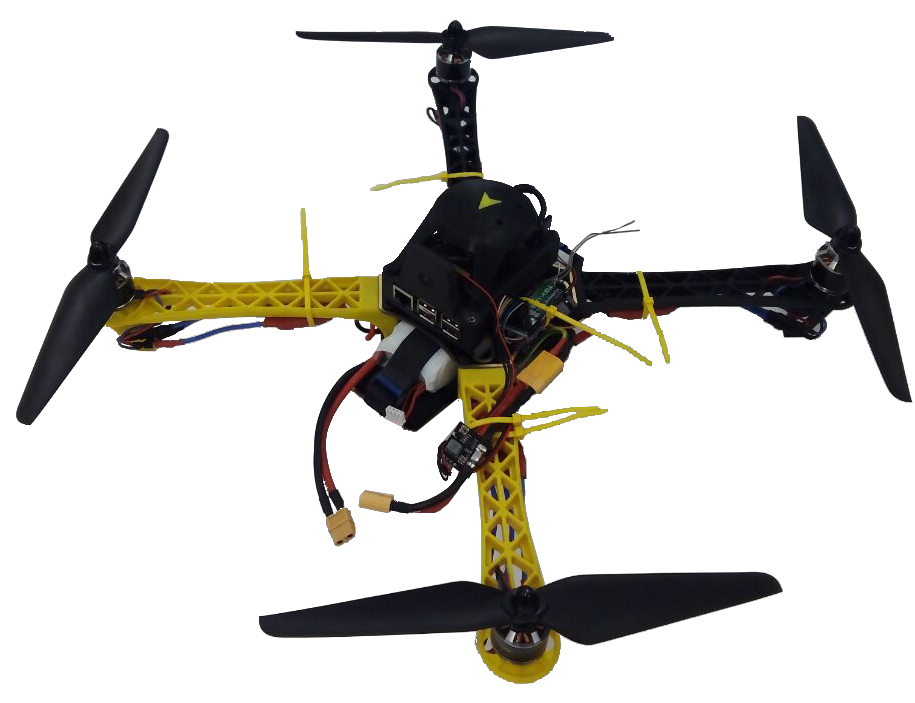}
\caption{Assembled Erle quadcopter}
\label{fig:erle}
\end{figure}

The performance of multi-robot systems is highly dependent on the platforms chosen, specially their sensing, actuating and communication capabilities, as well as how programmable they are.
In this work the platform chosen was the Erlecopter. This choice was made based on their low cost, ability for autonomous navigation, natively running ardupilot and compatibility with ROS. Figure \ref{fig:erle} shows the one of these quadcopters. These quadcopters, already possess a series of hardware characteristics which are deemed necessary for swarm application using real platforms, namely:

\begin{itemize}
\item On-board 1.2 GHz quad-core ARM cortex-53 CPU, 1GB RAM and Broadcom VideoCore IV.
\item Gravity Sensor.
\item Gyroscope.
\item Digital Compass.
\item Pressure and temperature sensor.
\item WiFi module (802.11n, 2.4GHz), compatible with ad-hoc networking
\item 8MP Fixed focus lens, 2592 x 1944 pixel static images, supports 1080p30, 720p60 and 640x480p60/90 video record
\end{itemize}


\section{Results}

\begin{figure}[H]
\centering
\includegraphics[width=\linewidth]{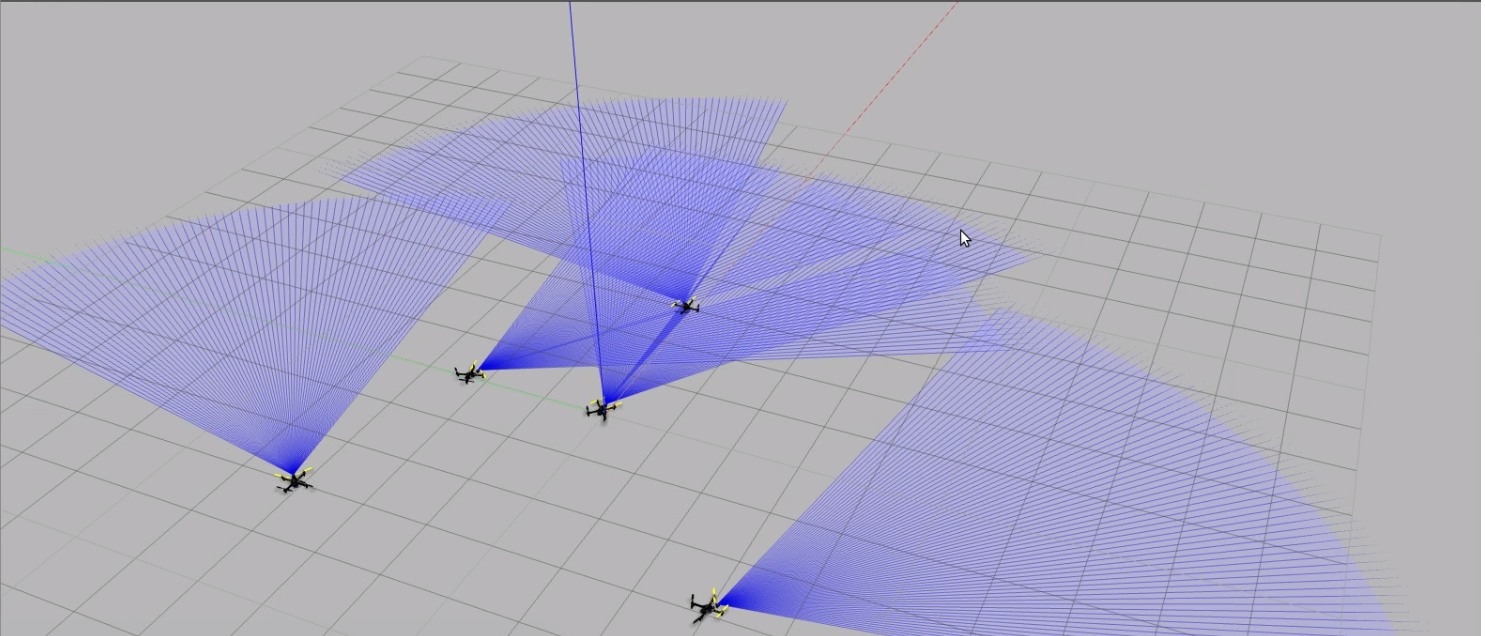}
\caption{Initial swarm configuration.}
\label{fig:initpos}
\end{figure}

 To demonstrate the potential of the changes made to ardupilot/gazebo plugin, we present below the values of the metrics used to assess flocking in our simulation. Figure\ref{fig:initpos} shows the initial position and orientation of the swarm.

\begin{figure}[!htp]
 \centering
        \includegraphics[width=1\linewidth]{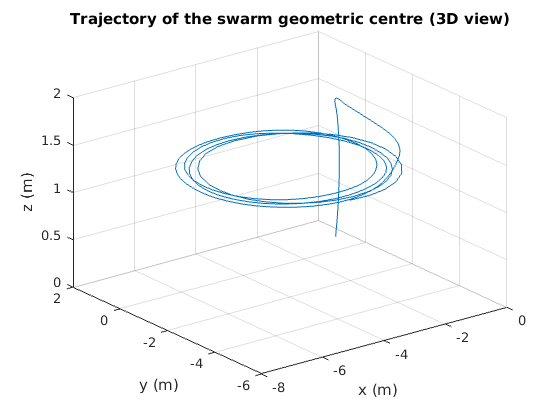}
        \caption{3D Visualization of the swarm's geometric centre trajectory}
        \label{fig:traj}
        \vspace{-10pt}
\end{figure}

 To understand the plots of the metrics themselves it is useful to observe first which trajectory the swarm took during simulation. For this purpose the geometric center is plotted in Figure \ref{fig:traj}. Note that, the initial point in time is that of a lower altitude, when the the UAVs have still to takeoff. The first stage of the plot shows the increasing in altitude of the geometric centre.

\noindent As one can observe from Figure \ref{fig:traj}, over time, the trajectory stabilizes towards a circular shape. In fact, this is somewhat expected since each UAV has a neighborhood only limited by distance and not by angle. This means that neighbours which are behind the UAV also influence the computation of the flocking rules.
\\ 
Nevertheless, this is an important result since it shows that the swarm as whole is able of acquiring a coordinated behaviour (a recording of this simulation can be found here \footnotemark \footnotetext{\href{https://www.youtube.com/watch?v=hVe3ht-fOtw}{https://www.youtube.com/watch?v=hVe3ht-fOtw}}).

\begin{figure}[!htp]
\centering
\includegraphics[width=1\linewidth]{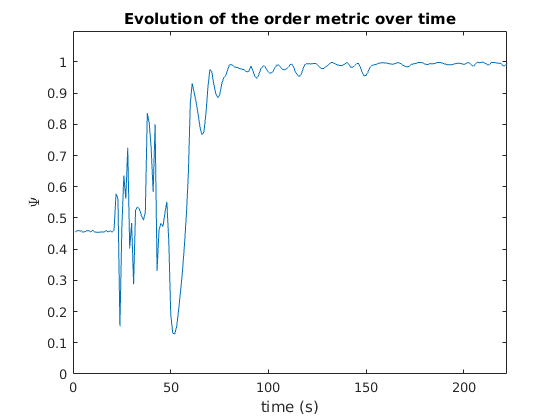}
\caption{Order metric over time}
\label{fig:order}
\end{figure}

 Furthermore, as it can be seen on Figure \ref{fig:order}, the \textit{order} value indeed tends to 1 over time. Naturally, in the beginning of the simulation it is very close to its minimum because velocities are initialized randomly. However,  over time it is clear the tendency of it to stabilize around 1. Note that this result corroborates the previous one. Since this metric measures the \textit{relative} orientation of the UAVs in the swarm, if the swarm makes a change of direction but their relative orientations remain similar, this metric would still yield a high value. Obviously some small valleys exist, which relate to the fact that some UAVs may start their turn slightly before the rest of the swarm. At this point we should highlight that orientation is calculated, not from the yaw angle (i.e. z-rotation) but from the $v_x$ and $v_y$ components. In fact, yaw control is still an open issue in the developers forum \footnotemark \footnotetext{\href{http://forum.erlerobotics.com/t/erlecopter-yaw-rotation-is-failling-in-gazebo/3200/6}{http://forum.erlerobotics.com/t/erlecopter-yaw-rotation-is-failling-in-gazebo/3200/6}}. 

\begin{figure}[H]
\centering
\includegraphics[width=1\linewidth]{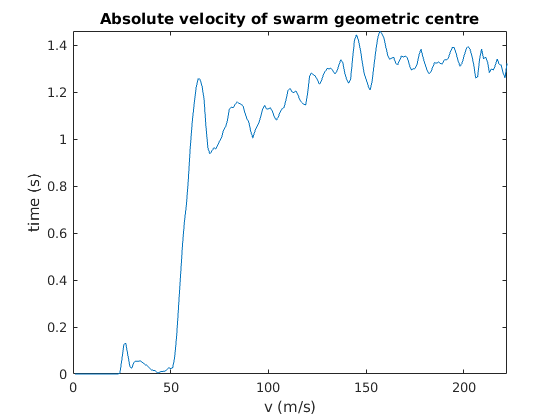}
\caption{Swarm centre's velocity over time}
\label{fig:sv}
\end{figure}

\noindent Finally, Figure \ref{fig:sv} depicts the value of the swarm's geometric centre velocity over time. It was mentioned previously that a desired swarm performance would translate into high value of this metric. In fact, if across the simulation, the UAVs would have opposite velocities, the resultant swarm velocity would have a very low norm. 
\vspace{-10pt}
\vspace{-10pt}
 The value of this metric increases quickly at approximately the same time as the order metric. This is also expected, specially if velocities have the same direction which in turn make the resultant velocity (i.e. $V_s$) increase as well. Lastly, this metric also shows a much more obvious evidence of the fact that the swarm stabilized in circular motion, seen in the consecutive increases and decreases of its value.\\
\vspace{-20pt}
\section{Discussion}

\noindent This work proposes a significant improvement on the functionalities of the Ardupilot plugin for Gazebo focusing on swarm robotics applications. 
The most relevant contribution is the clear opportunity it gives researchers to integrate open-source software, such as Ardupilot, into their own research. Moreover, as Ardupilot is becoming frequently used for off-the-shelf robots, it will also encourage research on the topic of swarm robotics and multi-robot systems. With the tools described in this paper a researcher does not have to develop their own platform and can focus on the algorithmic part of the implementation. Hence, the outcome of this work functions as an enabler of a vast number of swarm applications.








\end{document}